\documentclass[10pt,twocolumn,letterpaper]{article}

\usepackage{wacv}
\usepackage{times}
\usepackage{epsfig}
\usepackage{graphicx}
\usepackage{amsmath}
\usepackage{amssymb}
\usepackage{booktabs}
\usepackage{xcolor}
\usepackage{url}
\usepackage{array,multirow}

\def\wacvPaperID{33} 

\wacvapplicationstrack 

\wacvfinalcopy 

\usepackage{url}            
\usepackage{booktabs}       
\usepackage{amsfonts}       
\usepackage{nicefrac}       
\usepackage{microtype}      
\usepackage{xcolor}         
\usepackage{amsmath,graphicx}
\usepackage{subcaption}
\usepackage{amssymb}
\usepackage{xspace}
\usepackage{enumitem}
\usepackage[accsupp]{axessibility}

\ifwacvfinal
\usepackage[breaklinks=true,bookmarks=false]{hyperref}
\else
\usepackage[pagebackref=true,breaklinks=true,colorlinks,bookmarks=false]{hyperref}
\fi

\begin{document}

\title{Learning Robust Deep Visual Representations from EEG Brain Recordings \vspace{-5.0mm}}


\author{Prajwal Singh$^{\dagger}$\thanks{This work is supported by Prime Minister Research Fellowship (PMRF-2122-2557) and Jibaben Patel Chair in Artificial Intelligence.}~, Dwip Dalal$^{\dagger}$, Gautam Vashishtha$^{\dagger}$, Krishna Miyapuram$^{\ddagger}$, Shanmuganathan Raman$^{\dagger *}$\\
CVIG Lab$^{\dagger}$, BRAIN Lab$^{\ddagger}$\\
IIT Gandhinagar, India\\
{\tt\small \{singh\_prajwal, dwip.dalal, gautam.pv, kprasad, shanmuga\}@iitgn.ac.in}
}

\maketitle
\thispagestyle{empty}

\begin{abstract}
\label{ref:abstract}

Decoding the human brain has been a hallmark of neuroscientists and Artificial Intelligence researchers alike. Reconstruction of visual images from brain Electroencephalography (EEG) signals has garnered a lot of interest due to its applications in brain-computer interfacing. This study proposes a two-stage method where the first step is to obtain EEG-derived features for robust learning of deep representations and subsequently utilize the learned representation for image generation and classification. We demonstrate the generalizability of our feature extraction pipeline across three different datasets using deep-learning architectures with supervised and contrastive learning methods. We have performed the zero-shot EEG classification task to support the generalizability claim further. We observed that a subject invariant linearly separable visual representation was learned using EEG data alone in an unimodal setting that gives better k-means accuracy as compared to a joint representation learning between EEG and images. Finally, we propose a novel framework to transform unseen images into the EEG space and reconstruct them with approximation, showcasing the potential for image reconstruction from EEG signals. Our proposed image synthesis method from EEG shows $62.9\%$ and $36.13\%$ inception score improvement on the EEGCVPR40 and the Thoughtviz datasets, which is better than state-of-the-art performance in GAN \footnote{\url{github.com/prajwalsingh/EEGStyleGAN-ADA}}.
\vspace{-2mm}
\end{abstract}
\section{Introduction}
\label{ref:introduction}

The field of Brain-Computer Interface (BCI) has witnessed a surge in interest and research due to its potential to revolutionize the way machines are controlled through human cognition. This rapidly advancing field holds promise for transforming various domains of human-computer interaction by leveraging our understanding of human brain activity \cite{green2011learning, guger1999prosthetic, muller2007control}. In this context, Electroencephalography (EEG) has emerged as a key method for recording brain activity and has garnered significant attention in the scientific community \cite{jasper1958report}. EEG offers several advantages as a non-invasive technique, providing dense temporal information about brain activity. Its practical applications span a wide range of domains, including the identification of hand motor movements \cite{xu2021classification}, neurorehabilitation \cite{bamdad2015application}, and even the decoding of speech from brain signals \cite{defossez2022decoding}. Notably, extracting visual information from EEG signals has been a long-standing research focus within the BCI field \cite{carlson2011high, mustafa2012eeg, shenoy2008human, wang2012combining}. Recent progress in methods for extracting visual information from EEG signals has paved the way for exciting possibilities, such as the synthesis of images using learned EEG representations \cite{jiang2020brain, Spampinato2016DeepLH, spampinato2017deep, tirupattur2018thoughtviz}. However, existing approaches in this domain have encountered limitations regarding the quality of synthesized images and reliance on label supervision.

\begin{figure}[!t]
\centering
\begin{minipage}[b]{0.95\linewidth}
    \begin{subfigure}[t]{1.0\linewidth}
        \includegraphics[width=1.0\linewidth]{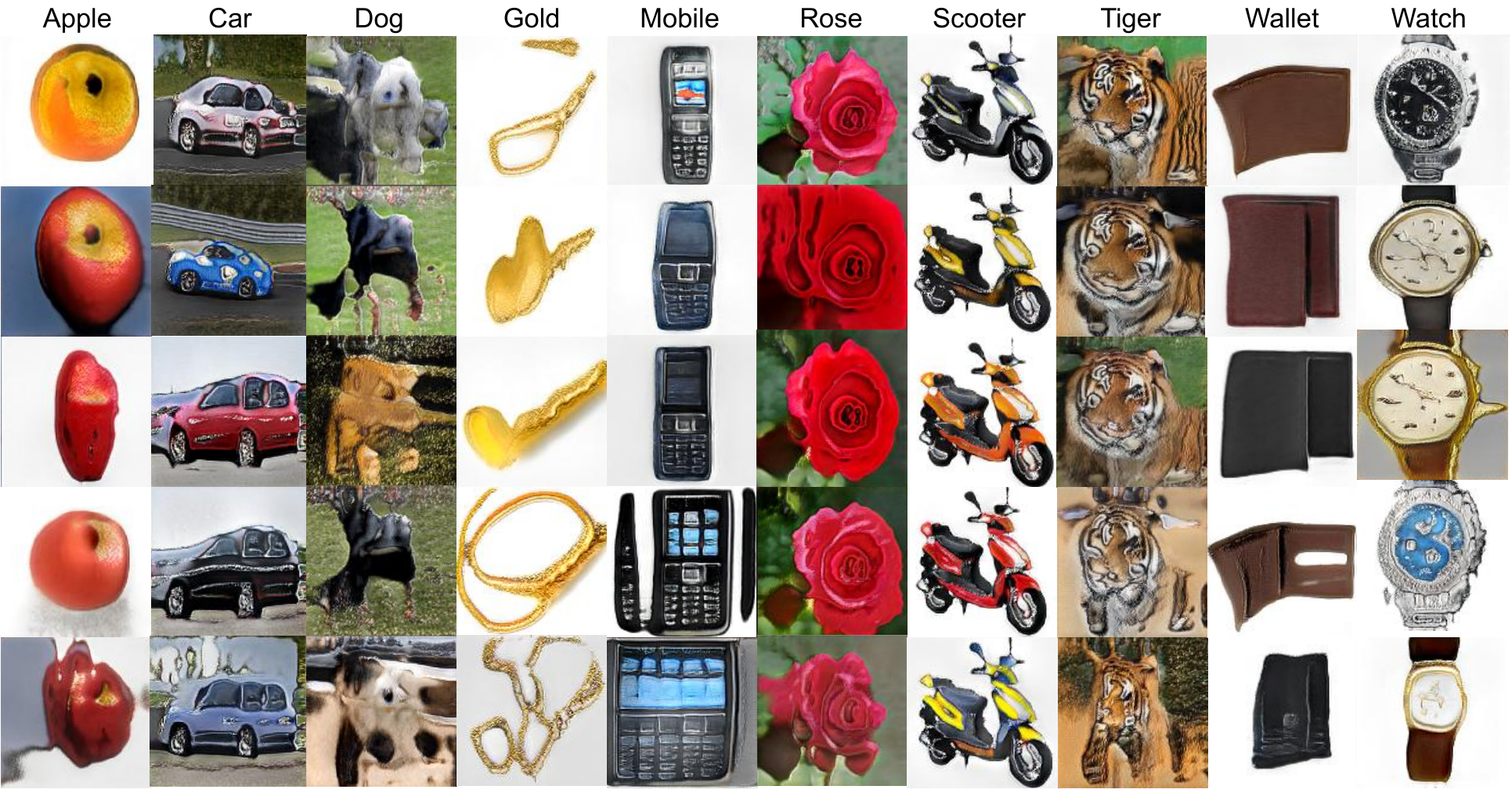}
    \end{subfigure}
\end{minipage}
\caption{\textbf{EEG to Image.} Sample images generated from EEGStyleGAN-ADA using EEG signals where each image is generated with different EEG signals across different classes, Thoughtviz dataset \cite{kumar2018envisioned, tirupattur2018thoughtviz}.}
\label{fig:genratedimages_thoughtviz}
\vspace{-4mm}
\end{figure}
\begin{figure*}[!t]
\centering
\begin{minipage}[b]{1.0\linewidth}
    \begin{subfigure}[t]{1.0\linewidth}
        \includegraphics[width=1.0\linewidth]{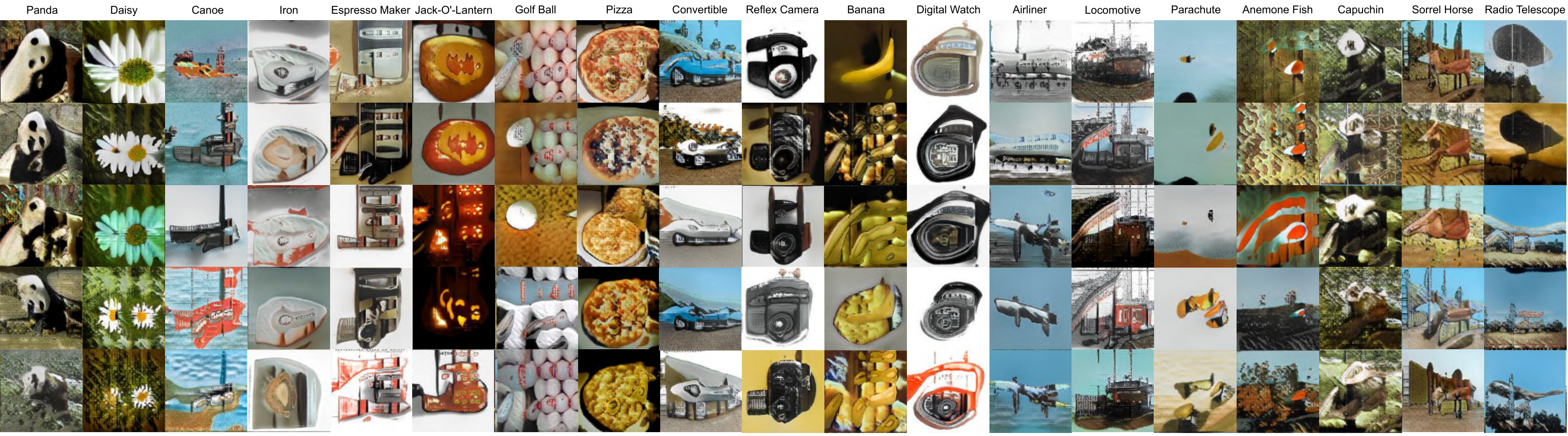}
    \end{subfigure}
\end{minipage}
\caption{\textbf{EEG to Image.} Sample images generated from EEGStyleGAN-ADA using EEG signals where each image is generated with different EEG signals across different classes, EEGCVPR40 dataset \cite{Spampinato2016DeepLH}.}
\label{fig:genratedimages}
\vspace{-4mm}
\end{figure*}

Our proposed work addresses these limitations and significantly advances EEG-based image synthesis.
\begin{enumerate}[noitemsep]
    \item We introduce an EEGStyleGAN-ADA framework, which improves image synthesis from EEG signals by leveraging learned EEG representations in contrastive settings. This approach generates higher-quality images, overcoming the drawbacks of previous methods and improving the state-of-the-art (SOTA) FID score by $62.9\%$ and $36.13\%$ on EEGCVPR40 \cite{Spampinato2016DeepLH} and ThoughtViz dataset \cite{kumar2018envisioned, tirupattur2018thoughtviz}.
    \item We investigate the impact of employing a similar architecture for feature extraction across all EEG datasets, thus reducing the overt dependency on the nature of the data distribution when modeling the architecture.
    \item To showcase the adaptability of the acquired representations from the proposed EEG feature extraction framework, we present results involving zero-shot classification performance and a novel image-to-image translation method designed to reconstruct previously unseen images directly from the EEG space.
    \item Our work further presents an innovative framework for joint representation learning that bridges the two different modalities, i.e., EEG and images, drawing inspiration from the existing Contrastive Language-Image Pre-Training (CLIP) method \cite{radford2021learning}. By fusing EEG signals and visual cues, our objective is to craft an enriched and comprehensive representation aimed to amplify performance across a spectrum of tasks, notably image classification and retrieval.
\end{enumerate}

We have also conducted several experiments and ablation studies exploring EEG feature extraction using supervised and metric learning-based methods across different architectures. This rigorous evaluation allows us to assess the effectiveness of our proposed approaches and shed light on the underlying mechanisms of EEG-based representation learning. Our work addresses the challenges faced in EEG-based image synthesis and representation learning tasks and offers novel frameworks and experimental insights. By improving the quality of the synthesized images, enabling joint representation learning, creating novel frameworks for EEG-based image reconstruction, and conducting comprehensive evaluations, we aim to push the boundaries of what is possible in harnessing EEG signals for visual tasks. 

\section{Related Works}
\label{ref:relatedworks}

In the past decade, deep learning-based methods have made it possible to learn representation from complex data such as EEG, images, or text. The initial work by \cite{Spampinato2016DeepLH} proposed an EEGCVPR40 \cite{Spampinato2016DeepLH} dataset along with an LSTM-based EEG classification network for feature learning. Following this, the work by Kavasidis and Palazzo \emph{et al.} \cite{kavasidis2017brain2image, palazzo2017generative} uses a Generative Adversarial Network (GAN) \cite{goodfellow2020generative} for synthesizing images from EEG features which learned using the LSTM based network proposed by \cite{Spampinato2016DeepLH}. Along with GAN, Kavasidis \emph{et al.} \cite{kavasidis2017brain2image} also used a Variational Autoencoder (VAE) \cite{kingma2013auto} for synthesizing images from EEG signals. Their work concluded that the GAN-based method outperform the VAE in synthesizing the photorealistic images. Tirupattur \emph{et al.} \cite{tirupattur2018thoughtviz} proposed a GAN network that learns from a small-size dataset \cite{kumar2018envisioned}. They have added a trainable Gaussian layer in the network that learns mean $\mu$ and variance $\sigma$ of the EEG feature, preventing the discriminator network from overfitting. The work by Mishra \emph{et al.} \cite{mishra2022neurogan} uses an attention-based GAN network along with a trainable Gaussian layer for synthesizing images from small-size EEG dataset \cite{kumar2018envisioned}. Both the work \cite{tirupattur2018thoughtviz, kumar2018envisioned} use the pre-trained image classification network for training the generator in GAN. In contrast, the work by Singh \emph{et al.} \cite{singh2023eeg2image} uses a metric learning-based approach for feature EEG extraction and modifies the GAN training strategy to use Differentiable Data Augmentation (DiffAug) \cite{zhao2020differentiable} method for overcoming the problem of small-size EEG dataset. This also reduces the network complexity, i.e., a trainable Gaussian layer and a pre-trained image encoder are not required for training the generator in GAN.

\begin{figure*}[!t]
\begin{minipage}[b]{0.10\linewidth}
  \centering
  \centerline{\includegraphics[width=\linewidth]{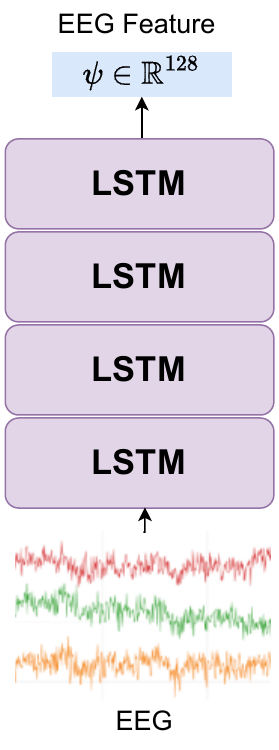}}
  \centerline{(a)}\medskip
  \vspace{-3mm}
\end{minipage}
\hfill
\begin{minipage}[b]{.13\linewidth}
  \centering
  \centerline{\includegraphics[width=\linewidth]{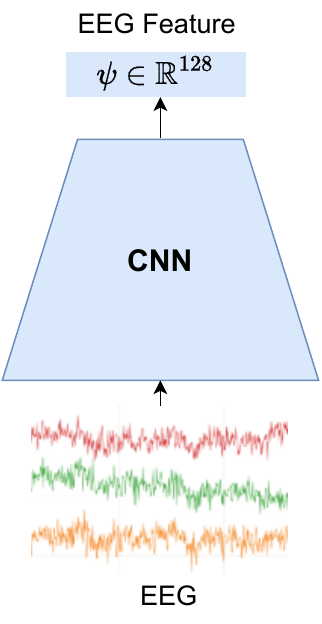}}
  \centerline{(b)}\medskip
 \vspace{-3mm}
\end{minipage}
\hfill
\begin{minipage}[b]{0.2\linewidth}
  \centering
  \centerline{\includegraphics[width=\linewidth]{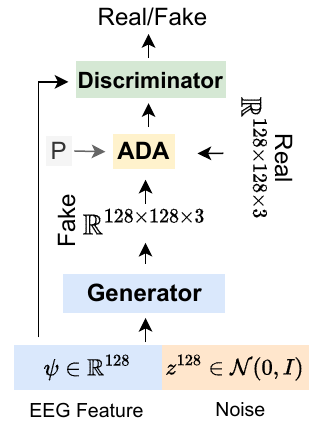}}
  \centerline{(c)}\medskip
 \vspace{-3mm}
\end{minipage}
\hfill
\begin{minipage}[b]{0.46\linewidth}
  \centering
  \centerline{\includegraphics[width=\linewidth]{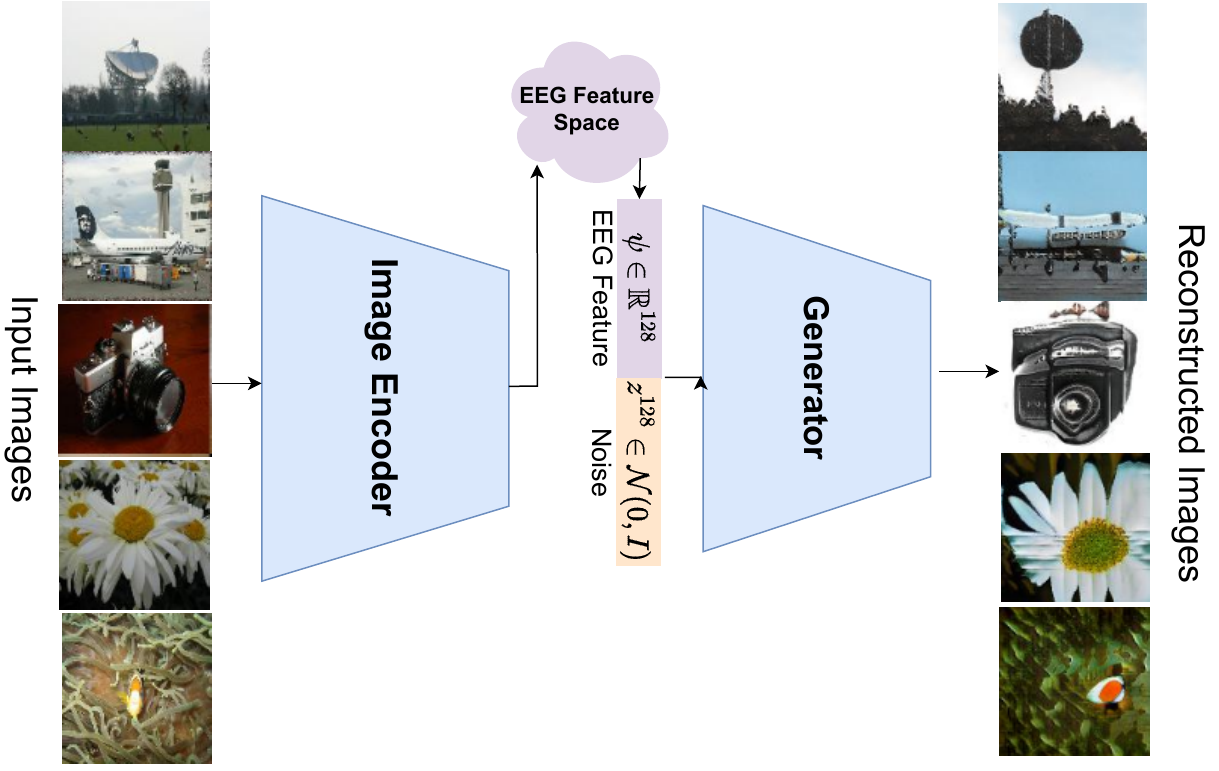}}
  \centerline{(d)}\medskip
 \vspace{-3mm}
\end{minipage}
\caption{(a) and (b) shows the LSTM and CNN architecture used for $128D$ feature extraction from EEG signal. (c) StyleGAN-ADA \cite{karras2020training} architecture with modification for EEG-based conditioning. (d) Illustrate the framework that transforms unseen images into learned EEG space, and then images are reconstructed from EEG features using the pre-trained generator network.}
\label{fig:architecture}
\vspace{-2mm}
\end{figure*}

\section{Method}
\label{ref:method}

This work aims to address the problem of EEG to image reconstruction by addressing three questions: 1) What are the different strategies for extracting visual information from the EEG data? 2) Can we reconstruct the images with rich information from extracted EEG features? and 3) How can we jointly train the EEG-Image model for tasks such as EEG-based image retrieval? To address these questions, we have performed experiments and ablation studies across three different datasets EEGCVPR40 \cite{Spampinato2016DeepLH}, ThoughtViz \cite{kumar2018envisioned, tirupattur2018thoughtviz} and Object \cite{kaneshiro2015representational} using different architectures with varying loss functions.

\subsection{Feature Extraction from EEG Data}

Extracting the features from the EEG data is an important step for problems like classification, reconstruction, or synthesizing images from EEG. Owing to the importance of feature extraction, several supervised or self-supervised methods have been used in the past. The following works \cite{Spampinato2016DeepLH, spampinato2017deep, zheng2020decoding, jiang2020brain, khare2022neurovision, zheng2020evoked}, use the supervised classification method for feature extraction. Supervised classification methods are preferable if the test data distribution overlaps with the train data distribution, which is not always the case with the EEG dataset. The issue can be overcome with self-supervised/metric-based learning and is addressed in these works \cite{jiang2019context, mukherjee2019cogni, mishra2021eeg, singh2023eeg2image}. In \cite{jiang2019context, mukherjee2019cogni}, a pre-trained image encoder is used to extract the features from the image, and an EEG encoder is trained to learn the feature distribution of images using regression and KL-divergence \cite{joyce2011kullback}. The work \cite{mishra2021eeg, singh2023eeg2image} uses the metric learning-based method for feature learning, where triplet loss \cite{schroff2015facenet} is used to train the EEG encoder.

\vspace{-10pt}

\begin{align}
    \scalebox{0.93}{$\displaystyle \min_{\theta}\mathbb{E}\big[ ||f_{\theta}(x^{a}) - f_{\theta}(x^{p})||_{2}^{2} - ||f_{\theta}(x^{a}) - f_{\theta}(x^{n})||_{2}^{2} + \delta \big]$}
    \label{eqn:1}
\end{align}

Where $f_{\theta}$ is our encoder and $x_{i} \in \mathbb{R}^{N \times C}$ is EEG input of $N$ time-steps and $C$ channels. In this work, we have also used the triplet loss for feature learning with semi-hard triplets. The semi-hard triplets prevent the encoder network from generating similar representations for all the data, and it enforces the learning of discriminative features. In Eq.\ref{eqn:1}, $x^{a}$ is a anchor, $x^{p}$ is a positive sample and $x^{n}$ is a negative sample. The $\delta$ is the margin distance between the positive and negative samples. The semi-hard triplets have the following property: $||f_{\theta}(x^{a}) - f_{\theta}(x^{p})|| < ||f_{\theta}(x^{a}) - f_{\theta}(x^{n})|| < (||f_{\theta}(x^{a}) - f_{\theta}(x^{p})|| + \delta)$.

\begin{figure}[!t]
\centering
\begin{minipage}[b]{0.85\linewidth}
    \begin{subfigure}[t]{1.0\linewidth}
        \includegraphics[width=1.0\linewidth]{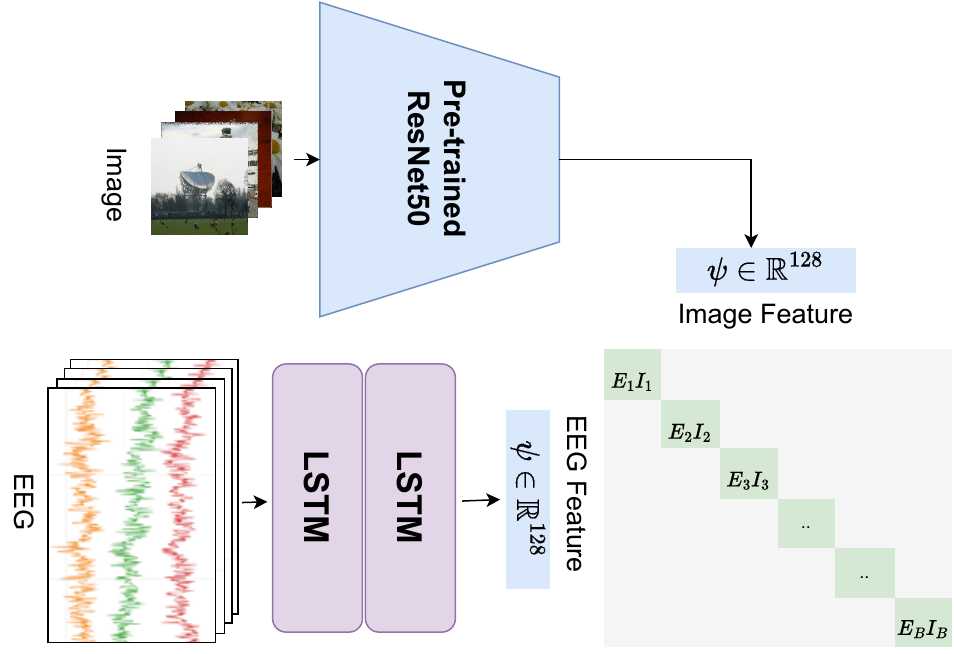}
    \end{subfigure}
\end{minipage}
\caption{\textbf{EEGClip.} Illustrate the architecture used for joint representation learning of EEG and image based on \cite{radford2021learning}.}
\label{fig:eegclip}
\vspace{-2mm}
\end{figure}

\begin{figure*}[!t]
\begin{minipage}[b]{0.33\linewidth}
  \centering
  \centerline{\includegraphics[width=\linewidth]{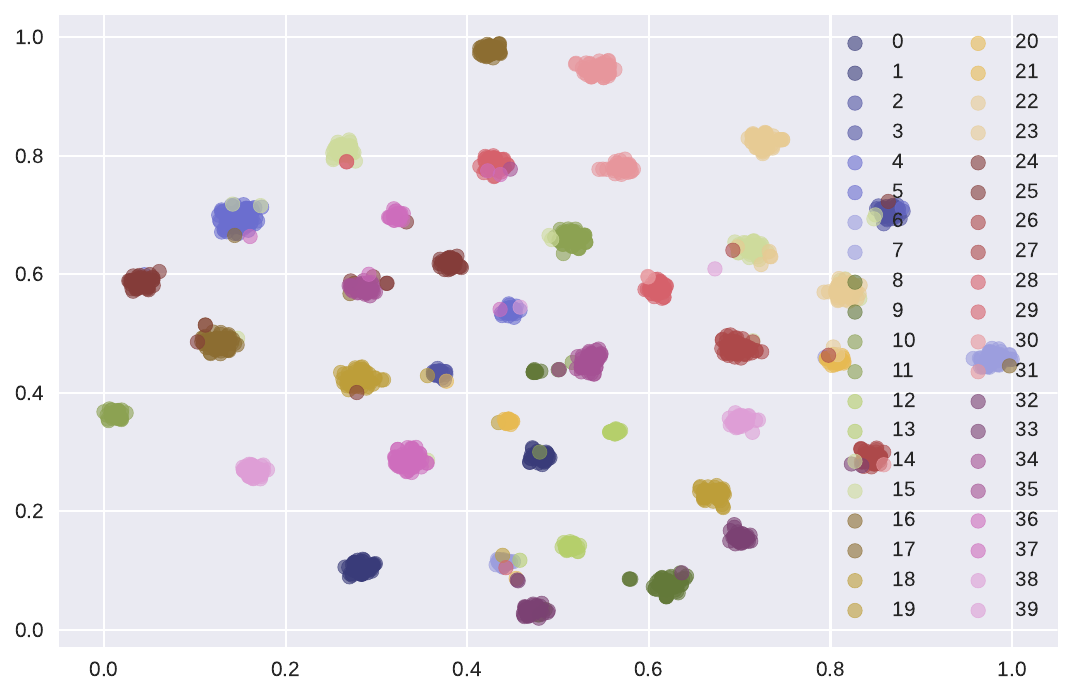}}
\end{minipage}
\begin{minipage}[b]{.33\linewidth}
  \centering
  \centerline{\includegraphics[width=\linewidth]{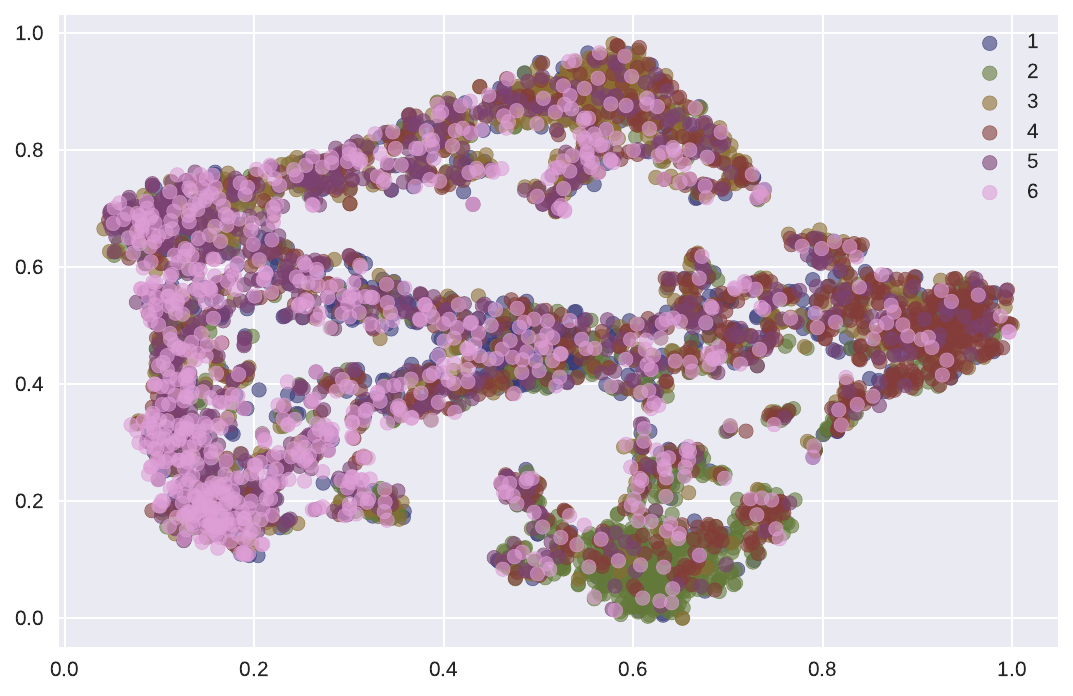}}
\end{minipage}
\begin{minipage}[b]{0.33\linewidth}
  \centering
  \centerline{\includegraphics[width=\linewidth]{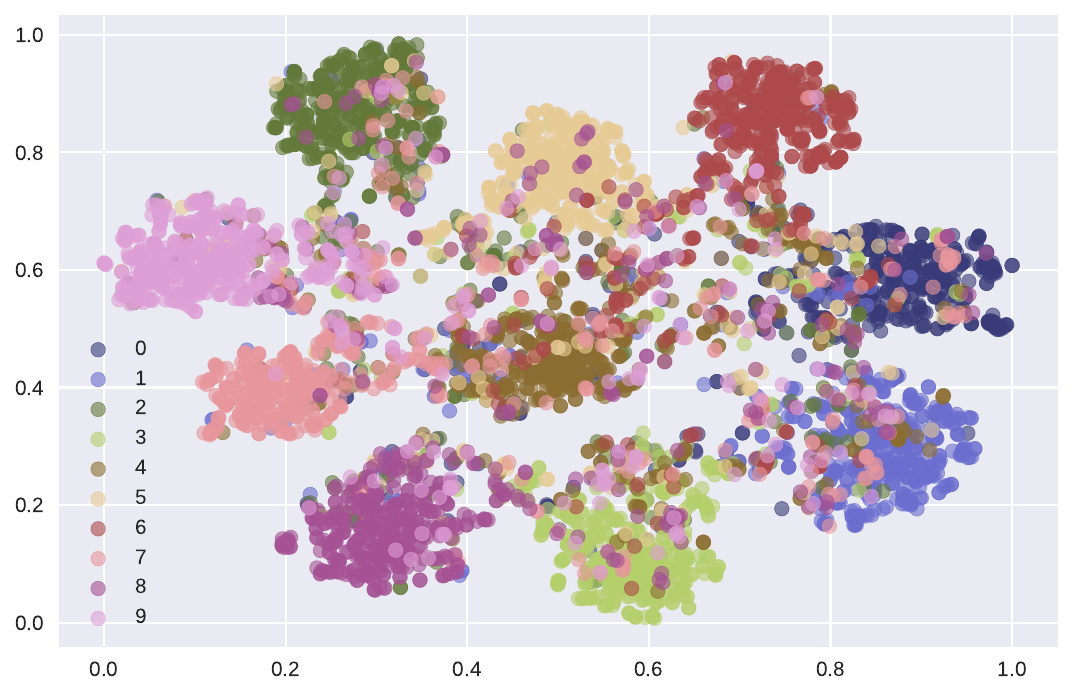}}
\end{minipage}
\begin{minipage}[b]{0.33\linewidth}
  \centering
  \centerline{\includegraphics[width=\linewidth]{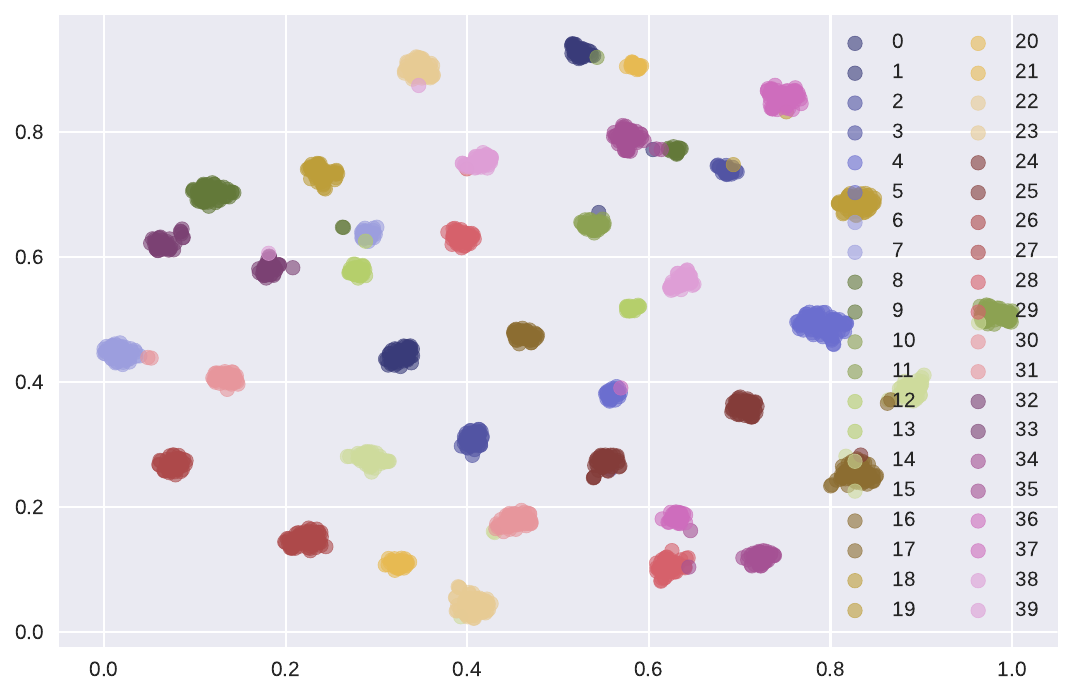}}
  \centerline{(a) EEGCVPR40 Dataset \cite{spampinato2017deep}}\medskip
  \vspace{-3mm}
\end{minipage}
\hfill
\begin{minipage}[b]{.33\linewidth}
  \centering
  \centerline{\includegraphics[width=\linewidth]{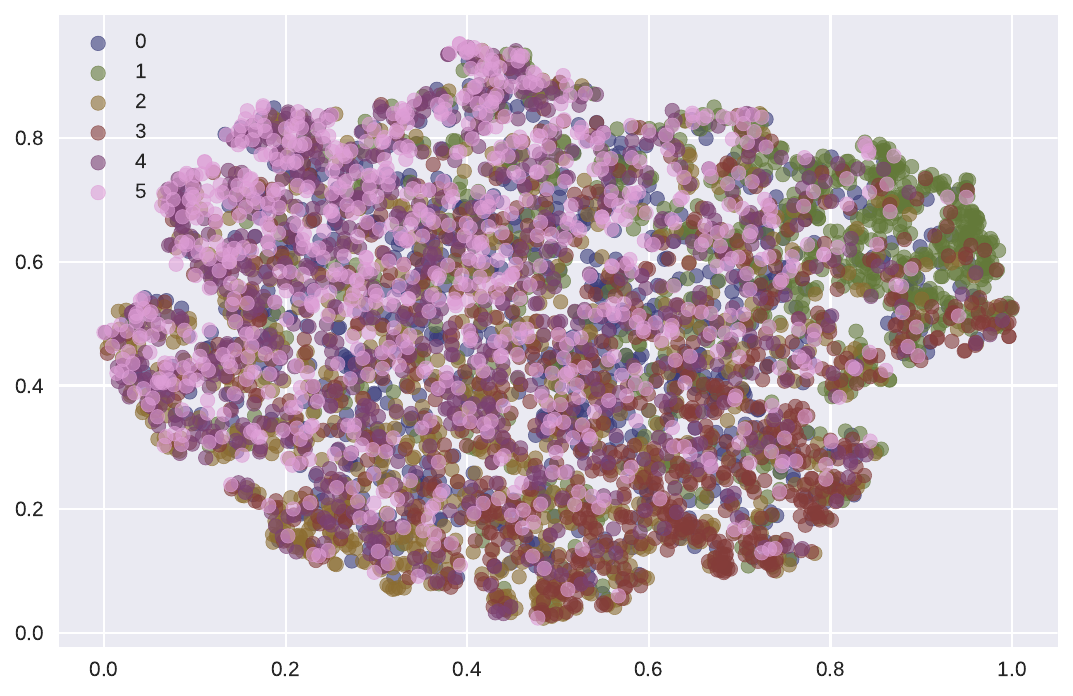}}
  \centerline{(b) Object Dataset \cite{kaneshiro2015representational}}\medskip
 \vspace{-3mm}
\end{minipage}
\hfill
\begin{minipage}[b]{0.33\linewidth}
  \centering
  \centerline{\includegraphics[width=\linewidth]{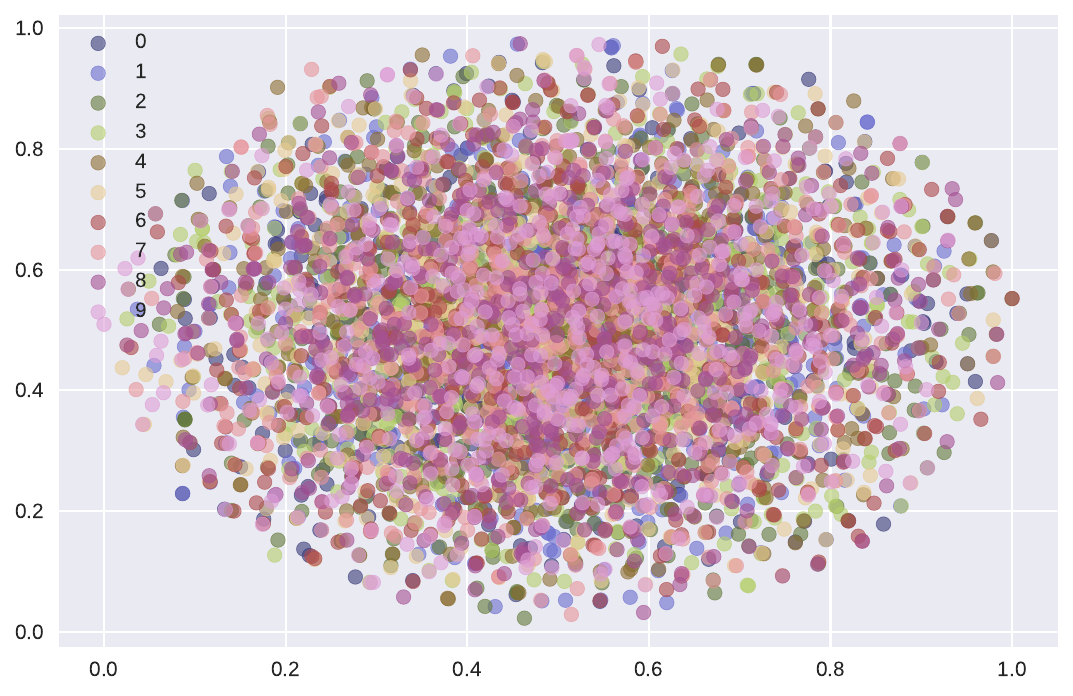}}
  \centerline{(c) ThoughtViz Dataset \cite{kumar2018envisioned, tirupattur2018thoughtviz}}\medskip
 \vspace{-3mm}
\end{minipage}
\caption{Figure illustrates EEG cluster learned with LSTM (top) and CNN (bottom) architecture using \underline{triplet loss}. K-Means score (a) $0.96, 0.98$ (b) $0.41, 0.35$ (c) $0.72, 0.12$. It shows that with a decrease in the timestep size of the EEG signal, the performance of CNN is also degrading. We kept the LSTM and CNN architecture the same across all the experiments to show the architecture's generalizability, which was not the case with previous methods where architectures were tailored according to the dataset.}
\label{fig:lstmcluster}
\vspace{-2mm}
\end{figure*}

\subsection{Generating Image from EEG Feature}

 The first work by Kavasidis and Palazzo \emph{et al.} \cite{palazzo2017generative, kavasidis2017brain2image} uses the GAN-based method to synthesize images using EEG features. Following this, \cite{khare2022neurovision, zheng2020decoding, fares2020brain} proposes modifying the GAN architecture to improve the image synthesis quality. In recent years, image synthesis with GAN has reached the limit of photorealistic images, which is indistinguishable from real images \cite{karras2020training, sauer2022stylegan}. Building on this, we proposed a framework for synthesizing images from EEG features using StyleGAN-ADA network \cite{karras2020training}. As shown in Fig.\ref{fig:architecture}, it takes feature vector and noise sampled from iso-tropic Gaussian distribution as input and synthesizes the desired image. StyleGAN-ADA \cite{karras2020training} uses adaptive discriminator augmentation, which helps the discriminator learn with limited data by augmenting real images at training time.

\subsection{Joint Space Learning for EEG and Image}

So far, the work has been using different networks for learning EEG and image representation space with the help of supervised and self-supervised methods. The modality of EEG signals and images is completely different, which makes learning the joint representation a non-trivial task. Few works have addressed the problem of joint representation learning between EEG and image. The work \cite{spampinato2017deep, jiang2019context, mukherjee2019cogni} uses a pre-trained image encoder to generate a representation for images equivalent to an EEG signal and train an EEG encoder network to regress the image feature vector. In work by Palazzo \emph{et al.} \cite{palazzo2020decoding}, they trained the EEG encoder in a contrastive setting with a triplet loss instead of regressing the image feature vector. This work utilizes the CLIP \cite{radford2021learning} based method for joint representation learning of EEG and image data. 

We used a pre-trained ResNet50 \cite{he2016deep} as an image encoder and a multilayer LSTM network as an EEG feature encoder. During training, we freeze the weight of ResNet50 and only update the weights of the LSTM network. We have used CLIP-based loss for training the complete pipeline. As shown in Fig.\ref{fig:eegclip}, each EEG-image pair is treated as a positive sample (diagonal elements) and the rest as a negative sample (non-diagonal elements). Similar to ours \cite{ye2022see} uses the CLIP \cite{radford2021learning} for joint representation learning, but their problem statement differs from ours where they aim to learn representation for image encoding by training GAN and later train the EEG encoder using the contrastive method for EEG based image retrieval. We have used a pre-trained image encoder for the EEG-based image retrieval tasks.
\section{Experiment and Results}
\label{ref:experiment}

In the first part of this section, we will discuss all the datasets used for training and testing. In the second part, we explain all the training regimes used for EEG feature learning and StyleGAN-ADA \cite{karras2020training} along with ablation studies. The later part of this section discusses the joint space representation learning EEG-Image CLIP \cite{radford2021learning} model.

\subsection{Datasets} 

We have used three datasets for training and testing the EEG representation learning.

\textbf{EEGCVPR40 Dataset} \cite{Spampinato2016DeepLH}. The dataset consists of EEG-Image pair for $40$ classes, which is a subset of the ImageNet \cite{deng2009imagenet} dataset. While recording the brain activity EEG signals, participants were shown $50$ images from each class for $0.5$ seconds. The EEG device consists of $128$ channels, and after preprocessing, the length of each EEG signal becomes $440$ timesteps. Approximately $11800$ EEG-Image pairs are there in the final dataset after discarding the bad samples.

\textbf{ThoughtViz Dataset} \cite{kumar2018envisioned, tirupattur2018thoughtviz}. This is a small-scale dataset curated by Kumar \emph{et al.} \cite{kumar2018envisioned}. It consists of $10$ different object classes, which is a subset of the ImageNet \cite{deng2009imagenet}. To collect the dataset, each participant was asked to visualize one of these $10$ different classes. The EEG device is of $14$ channels, and a total of 23 participants' brain activity was recorded. After pre-processing, each EEG signal becomes $32$ timesteps. In this work, we give it the alias ThoughtViz because it was first used by Tirupattur \emph{et al.} \cite{tirupattur2018thoughtviz} for EEG to image synthesis work and to avoid confusion with other Object \cite{kaneshiro2015representational} dataset.

\textbf{Object Dataset} \cite{kaneshiro2015representational}. The dataset consists of $6$ classes, each with 12 images that are shown to $10$ participants, and EEG signals are recorded using a $128$ channel device. The $6$ classes include a human body (HB), human face (HF), animal body (AB), animal face (AF), fruit vegetable (FV), and an inanimate object (IO). After pre-processing, the final EEG data is of size $124 \times 32$.

\begin{figure}[!t]
\centering
\begin{minipage}[b]{0.85\linewidth}
    \begin{subfigure}[t]{1.0\linewidth}
        \includegraphics[width=1.0\linewidth]{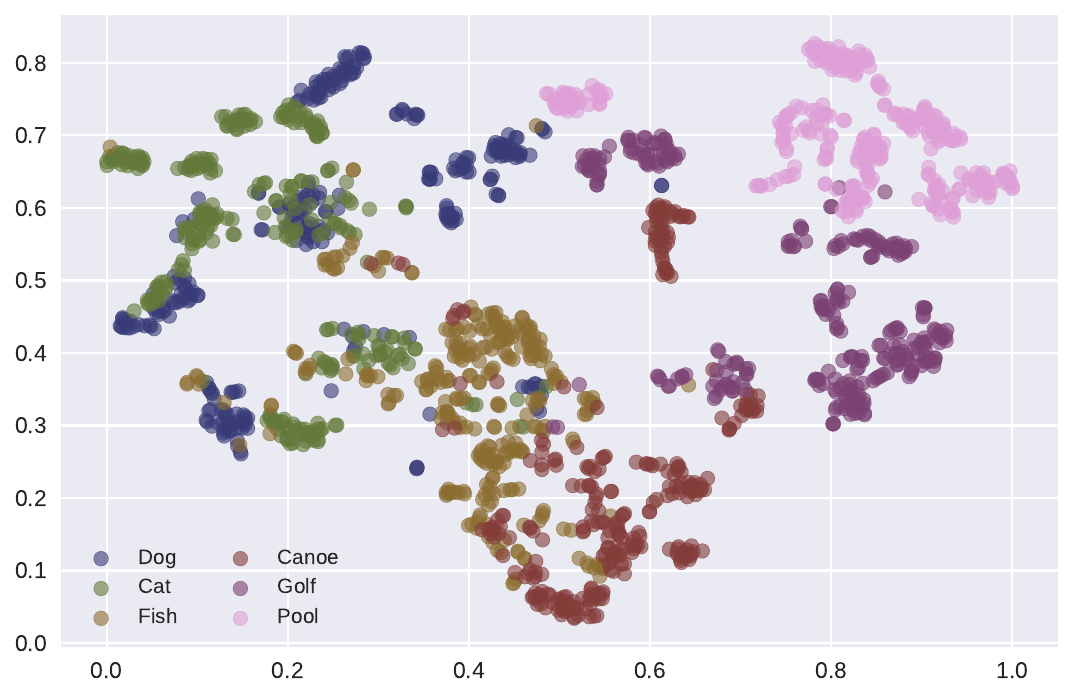}
    \end{subfigure}
\end{minipage}
\caption{\textbf{Unseen EEG Clustering.} Shows a t-SNE plot of 6 unseen categories features with a k-means accuracy of 0.62. It also shows the zero-shot classification generalizability of EEG features using the proposed method.}
\label{fig:unseeneegcluster}
\end{figure}

\begin{table}[!t]
\centering
\begin{tabular}{llll} \hline
\multicolumn{1}{c}{Method} & \multicolumn{1}{c}{SVM} & \multicolumn{1}{c}{kNN} & \multicolumn{1}{c}{K-Means} \\ \hline
Cogni-Net \cite{mukherjee2019cogni}             &  0.78  & 0.725                   & -                           \\
EEGLSTM (ours)              &   \textbf{0.93}  & \textbf{0.86}                   & 0.625                      \\ \hline
\end{tabular}
\caption{\textbf{Unseen EEG}. Shows the feature generalization capability of a network when trained using triple-loss. In this case, the network is trained on $34$ classes and tested $6$ unseen classes of EEGCVPR40 \cite{spampinato2017deep} dataset.}
\label{table:eegunseenclass}
\vspace{-4mm}
\end{table}

\subsection{Extracting EEG Features}

\textbf{Supervised.} We first train the feature extraction network using a supervised setting, where we use the label associated with each signal to penalize the network if it makes a wrong prediction. We have used two different architectures, as shown in Fig.\ref{fig:architecture}(a,b). We train both CNN and LSTM networks for all three datasets with label supervision. We observe that networks trained in supervision have very low k-means accuracy, which shows the learned features are not linearly separable except for the CNN network trained on the EEGCVPR40 \cite{Spampinato2016DeepLH} dataset.

\begin{table}[!t]
\resizebox{\linewidth}{!}{\begin{tabular}{lllll} \hline
Dataset                     & Method             & Accuracy & K-Means     & SVM         \\ \hline
\multirow{6}{*}{EEGCVPR40 \cite{Spampinato2016DeepLH}}  & LSTM Encoder \cite{spampinato2017deep}       & 0.829    & 0.45           & 0.47           \\
                            & DML \cite{jiang2019context}               & 0.977    & -           & -           \\
                            & LSTM-CNN \cite{zheng2020decoding}          & 0.944    & -           & -           \\
                            & BioLSTM \cite{jiang2020brain}           & 0.991    & -           & -           \\
                            & NeuroVision \cite{khare2022neurovision}       & 0.988    & -           & -           \\
                            & EEGLSTM (ours)          & \textbf{0.983}        & 0.961  & 0.962 \\ \hline
\multirow{2}{*}{Object \cite{kaneshiro2015representational}}     & BioLSTM \cite{jiang2020brain}           & 0.611    & -           & -           \\
    & ERP-LSTM \cite{zheng2020evoked}   & 0.66 & - & - \\
                            & EEGLSTM (ours) & \underline{0.41}        & 0.40   & 0.401 \\ \hline
\multirow{4}{*}{ThoughtViz \cite{kumar2018envisioned, tirupattur2018thoughtviz}} & ThoughtViz \cite{tirupattur2018thoughtviz}        & 0.729    & 0.18        & 0.19        \\
                            & SiameseCNN \cite{mishra2021eeg}        & 0.899    & -           & -           \\
                            & EEG2Image  \cite{singh2023eeg2image}        & 0.55        & 0.52        & -           \\
                            & EEGLSTM (ours) & \textbf{0.741}       & 0.721       & 0.724      \\ \hline
\end{tabular}}
\caption{\textbf{Clustering and Linear evaluation}. Comparison of different methods and loss types used for feature extraction from EEG signals across different datasets. In some cases, the triplet loss-based method outperforms the network trained with label supervision.}
\label{table:eegclustering}
\vspace{-4mm}
\end{table}

\textbf{Triplet Loss.} In this regime, we train both LSTM and CNN networks shown in Fig.\ref{fig:architecture}(a,b) using triplet loss \cite{schroff2015facenet} for all three datasets. Training networks using triplet loss helps them learn discriminative features, which leads to better k-means accuracy, as shown in Table \ref{table:eegclustering}. We have also shown the t-SNE \cite{van2008visualizing} plot of the learned representation of all three datasets in Fig.\ref{fig:lstmcluster} using the LSTM and CNN network. We further finetuned the network on all three datasets and reported the accuracy for comparison with other methods. It is vital to note that in order to show generalization in the EEG feature extraction task similar to using ResNet50 \cite{he2016deep} for image feature extraction across various datasets and applications, we have used the same LSTM/CNN architecture across all the EEG datasets. This explains the reason for having lower fine-tuning accuracy in some datasets. To further support this claim, for EEGCVPR40 \cite{Spampinato2016DeepLH} dataset under the same training regime, it has k-means accuracy of $98\%$ for CNN and $96\%$ for LSTM architectures.

\textbf{Unseen Data.} In order to show the generalizability of our learned representations using triplet loss across unseen classes, we have compared our method with \cite{mukherjee2019cogni} for this. In this regime, the network is trained on $34$ classes from EEGCVPR40 \cite{Spampinato2016DeepLH} dataset and tested on the remaining $6$ classes, which are a dog, cat, fish, canoe, golf, and pool. Compared to \cite{mukherjee2019cogni}, a pre-trained image network is not required. As shown in Table \ref{table:eegunseenclass}, our method performs better and has a higher SVM \cite{708428} and kNN \cite{Mucherino2009} score. We have also shown t-SNE \cite{van2008visualizing} plot for all the $6$ unseen classes learned features Fig.\ref{fig:unseeneegcluster}.

\begin{figure}[!t]
\centering
\begin{minipage}[b]{1.0\linewidth}
    \begin{subfigure}[t]{1.0\linewidth}
        \includegraphics[width=1.0\linewidth]{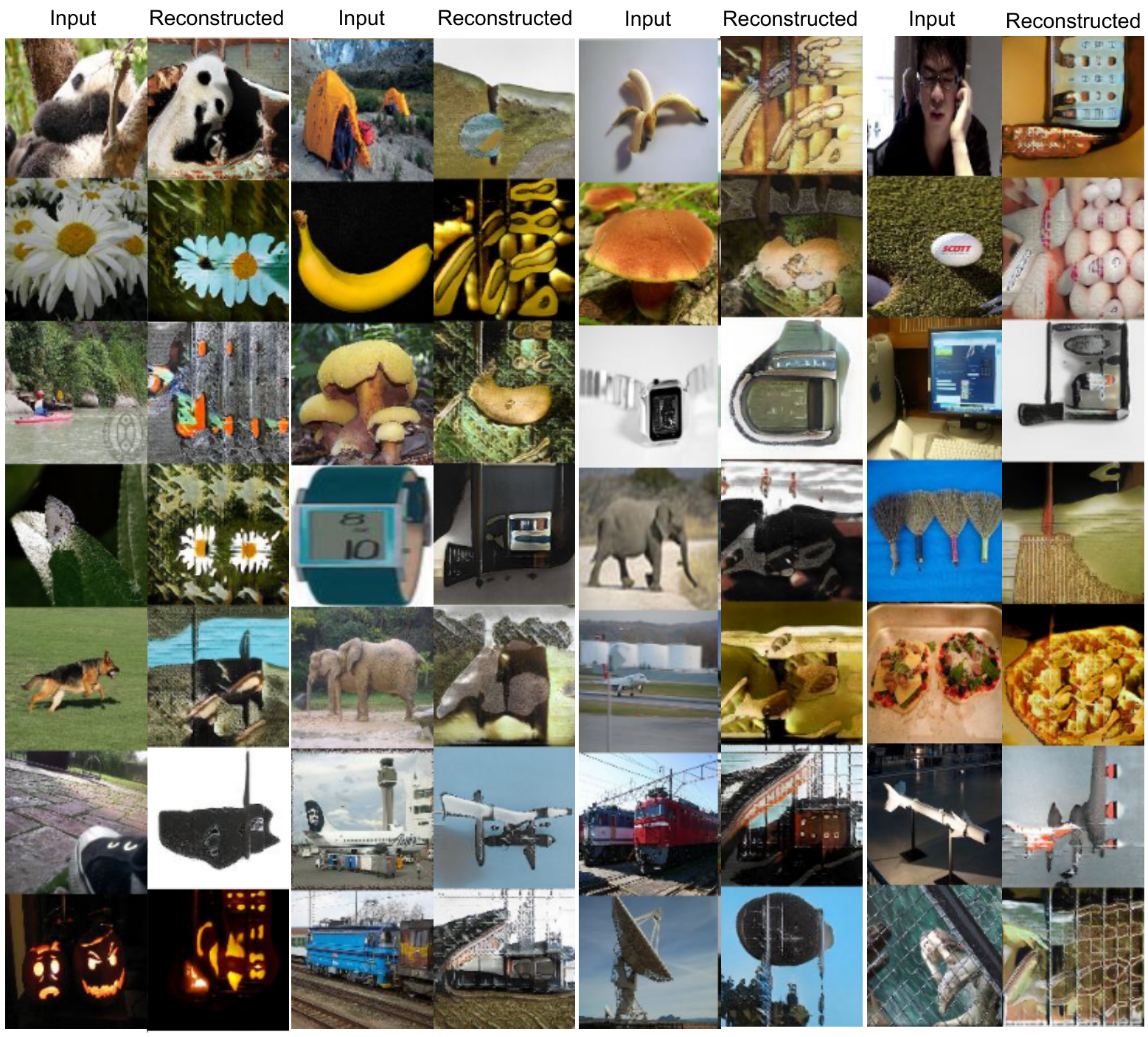}
    \end{subfigure}
\end{minipage}
\caption{\textbf{Image to Image.} Shows the result of Image to Image translation. Here, instead of using an EEG signal from EEGCVPR40 \cite{Spampinato2016DeepLH} dataset, its equivalent images are used, which are transformed into EEG representation space, and later, the image is reconstructed with approximation using a pre-trained generative network.}
\label{fig:image2image}
\vspace{-4mm}
\end{figure}

\textbf{Image to Image.} To study the effect of feature generalization, we performed another experiment, mapping the visual image features into the learned EEG manifold. We have learned the EEG feature space using triplet loss. The concept of mapping visual features stems from the notion that these mapped features can emulate the neural processes involved in human scene understanding \cite{Spampinato2016DeepLH}. These transformed image features were then used for image synthesis using EEGStyleGAN-ADA to show the generalization ability of the proposed network. The qualitative result of image synthesis is shown in Fig.\ref{fig:image2image}. The EEG features generated from unseen images can reconstruct the images with high fidelity.

\begin{table}[!t]
\centering
\resizebox{0.46\textwidth}{!}{%
\begin{tabular}{lllll}
\hline
 & \multicolumn{1}{l}{Method} & \multicolumn{1}{l}{IS $\uparrow$} & \multicolumn{1}{l}{FID $\downarrow$} & \multicolumn{1}{l}{KID $\downarrow$} \\
\hline
\multirow{5}{*}{\rotatebox[origin=r]{90}{EEGCVPR40}} & Brain2Image-VAE \cite{kavasidis2017brain2image} & 4.49 & - & - \\
& Brain2Image-GAN \cite{palazzo2017generative, kavasidis2017brain2image} & 5.07 & - & - \\
& NeuroVision \cite{khare2022neurovision} & 5.15 & - & - \\
& Improved-SNGAN \cite{zheng2020decoding} & 5.53 & - & - \\
& DCLS-GAN \cite{fares2020brain} & 6.64 & - & - \\
& EEGStyleGAN-ADA (ours) & \textbf{10.82} & 174.13 & 0.065 \\
\hline
\multirow{5}{*}{\rotatebox[origin=c]{90}{ThoughtViz}} & AC-GAN \cite{odena2017conditional} & 4.93 & - & - \\
& ThoughtViz \cite{tirupattur2018thoughtviz} & 5.43 & - & - \\
& NeuroGAN \cite{mishra2022neurogan} & 6.02 & - & - \\
& EEG2Image \cite{singh2023eeg2image} & 6.78 & - & - \\
& EEGStyleGAN-ADA (ours) & \textbf{9.23} & 109.49 & 0.039 \\
\hline        
\end{tabular}
}
\caption{Comparison of Inception Score (on all classes) of EEGCVPR40 \cite{Spampinato2016DeepLH} and ThoughtViz dataset \cite{kumar2018envisioned, tirupattur2018thoughtviz}. For EEGStyleGAN-ADA, we have also calculated Frechet Inception Distance (FID) and Kernel Inception Distance (KID).}
\vspace{-4mm}
\label{table: combined}
\end{table}

\begin{table*}[!t]
\centering
\resizebox{\textwidth}{!}{\begin{tabular}{cccccccc} \hline
\multicolumn{8}{c}{Fine Tuning Top-K Accuracy {[}@1 / @5 / @10{]}}                                                                                         \\
\multicolumn{2}{c}{Batch Size\textbackslash{}Epochs}       & 64                   & 128                  & 256                  & 512                           & 1024    & 2048             \\ \hline
\multirow{3}{*}{EEG} & 16                                 & 0.26/0.45/0.60 & 0.37/0.59/0.71 & 0.51/0.75/0.81  & 0.59/0.80/0.85          &  0.69/0.88/0.92  & 0.73/0.90/0.93  \\
& 32                                & 0.32/0.53/0.68  & 0.43/0.68/0.79 & 0.53/0.80/0.87 & 0.61/0.87/0.90          &  0.72/0.92/0.95  & 0.77/0.94/0.96   \\
& 64                                & 0.34/0.54/0.69 & 0.44/0.67/0.82 & 0.57/0.80/0.85 & 0.68/0.89/0.93 &  0.76/0.94/0.97 & 0.79/0.96/0.98 \\ \hline 
\multirow{3}{*}{Image} & 16                                 & 0.78/0.88/0.91 & 0.83/0.90/0.92 & 0.87/0.93/0.95  & 0.89/0.96/0.98          &  0.91/0.97/0.98    & 0.92/0.97/0.99 \\
& 32                                & 0.80/0.91/0.94  & 0.84/0.92/0.95 & 0.90/0.96/0.98 & 0.92/0.97/0.99          &  0.92/0.96/0.98      & 0.93/0.96/0.99    \\
& 64                                & 0.84/0.94/0.95 & 0.88/0.95/0.98 & 0.91/0.97/0.98 & 0.93/0.98/0.99 &  0.94/0.97/0.99      & 0.95/0.99/1.0  \\ \hline 
\end{tabular}}
\caption{\textbf{EEGClip.} Results of EEGClip network finetuned on EEGCVPR40 \cite{Spampinato2016DeepLH} dataset for both EEG and image classification.}
\label{table:clip_eeg}
\vspace{-2mm}
\end{table*}

\subsection{Image Synthesis}
\label{sec:imagesynthesis}

\textbf{EEG Based Conditioning.} We have significantly improved EEG to image generation by training StyleGAN-ADA \cite{karras2020training}, a state-of-the-art generative model. To tailor it specifically for EEG data, we introduced modifications to the existing StyleGAN-ADA pipeline, resulting in a framework we named EEGStyleGAN-ADA (Fig. \ref{fig:architecture}(c)). Our approach incorporates a pre-trained LSTM network with triplet loss to extract EEG features, which are concatenated with the noise vector sampled from an isotropic Gaussian distribution. This combined input is then fed into the EEGStyleGAN-ADA network for image synthesis. For training the network, we employed the 'cifar' hyperparameters, leveraging data available from the EEGCVPR40 \cite{Spampinato2016DeepLH} and ThoughtViz \cite{kumar2018envisioned, tirupattur2018thoughtviz} datasets. The synthesized images produced by our proposed framework, as depicted in Fig. [\ref{fig:genratedimages_thoughtviz}, \ref{fig:genratedimages}], exhibit diversity and maintain a high level of fidelity when compared to previous methods. 

To quantitatively evaluate the performance of our approach, we employed the Inception score \cite{salimans2016improved}, a commonly used metric in generative models. Comparative analysis with existing EEG to image synthesis networks, as summarized in Table \ref{table: combined}, reveals that our proposed method outperforms them in terms of the Inception score. Furthermore, we report the Frechet Inception Distance (FID) \cite{bynagari2019gans} and Kernel Inception Distance (KID) \cite{binkowski2018demystifying} scores, providing additional insights into the quality and diversity of the generated images.

\begin{figure}[!t]
\centering
\begin{minipage}[b]{1.0\linewidth}
    \begin{subfigure}[t]{1.0\linewidth}
        \includegraphics[width=1.0\linewidth]{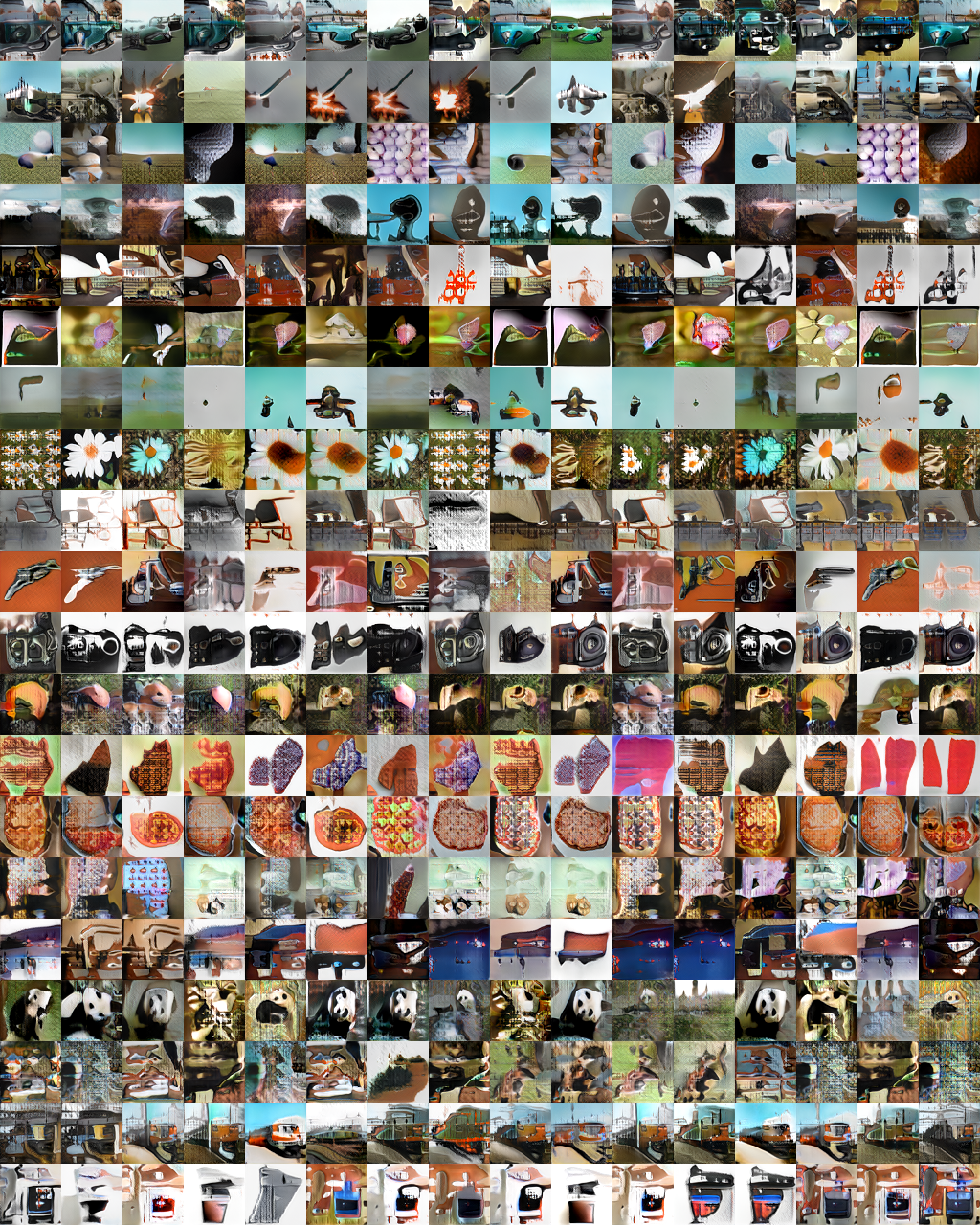}
    \end{subfigure}
\end{minipage}
\caption{\textbf{Class Based Conditioning.} To show the complexity of generating images using EEG from EEGCVPR40 \cite{Spampinato2016DeepLH} dataset. We train a variation of StyleGAN2 known as NoisyTwins \cite{rangwani2023noisytwins} for learning the image generation on long-tail conditional datasets or conditional datasets with fewer images, which is the case with the EEGCVPR40 dataset. Instead of using EEG, we have used one-hot encoding for conditioning. The best FID score we achieved is 105.5 with class label condition.}
\label{fig:stylegan_ablation}
\vspace{-4mm}
\end{figure}

\textbf{Class Based Conditioning.} To show the effectiveness of the proposed EEGStyleGAN-ADA, we performed an ablation study where instead of giving an EEG signal, we used one-hot class conditions only. The EEGCVPR40 \cite{Spampinato2016DeepLH} dataset consists of $40$ classes, each with $30$-$40$ images, making it difficult to learn using a conditional GAN. To further verify this claim, we perform the experiment with the current state-of-the-art, NoisyTwins \cite{rangwani2023noisytwins}, for the long-tail conditional generation. As shown in Fig.\ref{fig:stylegan_ablation}, the best FID score we achieved is $105.5$, and the qualitative result shows even after using the SOTA model for image synthesis using one-hot class conditioning on the EEGCVPR dataset lacks photorealistic effect. This implies that the photorealistic images we are synthesizing with the proposed EEGStyleGAN-ADA are the best among all the GAN state-of-the-art methods for EEG-based image generation.

\subsection{Joint Representation Space Learning}
\label{sec:jointrepspace}

\begin{figure}[!t]
\centering
\begin{minipage}[b]{0.9\linewidth}
    \begin{subfigure}[t]{1.0\linewidth}
        \includegraphics[width=1.0\linewidth]{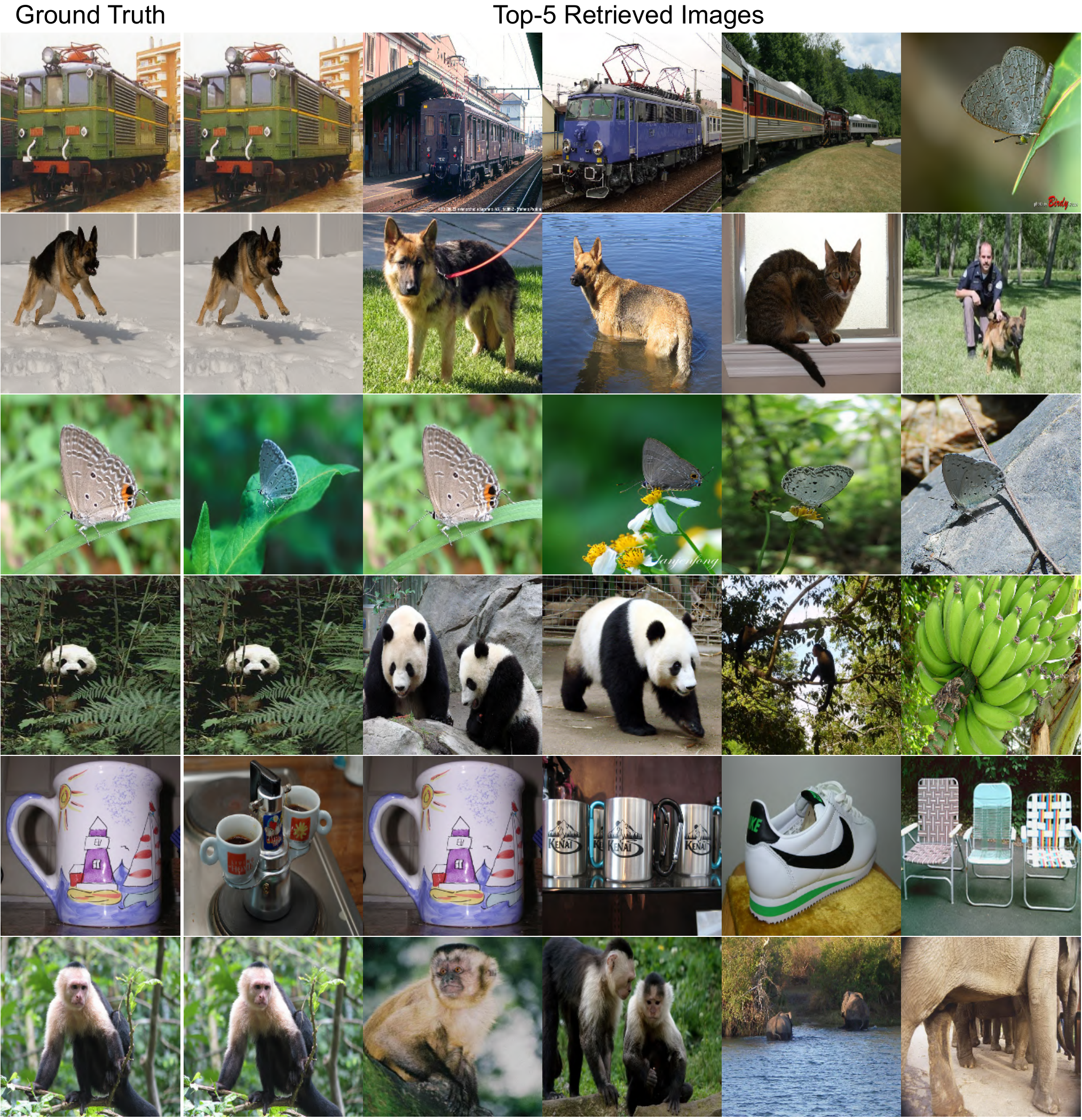}
    \end{subfigure}
\end{minipage}
\caption{\textbf{Image Retrieval using EEG.} Shows the top-$5$ retrieved images for the given EEG signal from test data of EEGCVPR40 \cite{Spampinato2016DeepLH}. The pre-trained weights of EEGClip are used to extract the feature for both the image and EEG. Here, ground truth shows the expected image equivalent to the given EEG.}
\label{fig:eegimageretrieval}
\vspace{-4mm}
\end{figure}

This section of our work presents EEGClip, a novel framework for joint representation learning between EEG signal and images built upon the CLIP Model \cite{radford2021learning}. To evaluate the effectiveness of our method, we conducted experiments using the EEGCVPR40 dataset \cite{Spampinato2016DeepLH}, which offers a significantly larger number of (EEG, Image) pairs compared to the ThoughtViz \cite{kumar2018envisioned, tirupattur2018thoughtviz} and Object \cite{kaneshiro2015representational} datasets. We performed several experiments to investigate the impact of batch size and the number of training epochs on learning joint representations. Due to computational constraints, we considered batch sizes of ${16, 32, 64}$ and trained the model for varying numbers of epochs, ranging from ${64, 128, 256, 512, 1024, 2048}$. The results of these experiments, as summarized in Table \ref{table:clip_eeg}, present the top-K recall rates for $K \in \{1, 5, 10\}$. Our findings indicate that the proposed EEGClip framework achieved superior performance when trained with a batch size of $64$ and for $2048$ epochs. This configuration yielded the highest recall rates for different values of $K$. To provide a visual representation of the effectiveness of EEGClip, we present image retrieval results from EEG in Fig.\ref{fig:eegimageretrieval}. These results demonstrate the ability of our framework to retrieve relevant images based on EEG input accurately.

\subsection{Experiment on EEGCVPR40 Filter Dataset (5-95 Hz)}

Based on the section \ref{sec:jointrepspace}, for EEGClip, the optimal performance is achieved when using a batch size of 64. Consequently, we employed the same batch size when conducting experiments on the EEGCVPR40 filter dataset (5-95 Hz) \cite{Spampinato2016DeepLH}. We have compared our method with  Palazzo \textit{et al.} \cite{palazzo2020decoding}, where they conducted experiments on the same filter dataset and achieved an accuracy of $60.4\%$ for EEG classification and $94\%$ for image classification. Our best-performing model, which corresponds to a batch size of $64$ and $2048$ epochs (as shown in Table \ref{table:clip_eegg}), achieved an accuracy of $64\%$ for EEG classification and $94\%$ for image classification.

\subsection{Quantitative Performance of Image Retrieval Task}

We employed the EEGClip model (Fig. \ref{fig:eegclip}) based on EEG inputs for the Image Retrieval task. The outcomes obtained from our experiments have been presented in the main paper. However, to provide a more comprehensive analysis of the results, we utilized two specific metrics: mean Reciprocal Rank (MRR) and mean Average Precision (mAP) for ranking evaluation. The MRR metric assesses the effectiveness of the retrieval model in accurately ranking the unique visually-cued instances, while the mAP metric evaluates the retrieval model's ability to capture all relevant visual cues. These relevant visual cues correspond to images that belong to the same semantic class as the correct match. For the EEGCVPR40 \cite{Spampinato2016DeepLH} dataset, our approach yielded an MRR of 0.7427 and an mAP of 0.6689. These scores were achieved using the model associated with a batch size of 64 and trained for 2048 epochs.

\begin{table*}[!t]
\centering
\resizebox{0.95\textwidth}{!}{
\begin{tabular}{cccccccc} \hline
\multicolumn{8}{c}{Fine Tuning Top-K Accuracy {[}@1 / @5 / @10{]}}                                                                                         \\
\multicolumn{2}{c}{Batch Size\textbackslash{}Epochs}       & 64                   & 128                  & 256                  & 512                           & 1024    & 2048             \\ \hline
EEG 
& 64 & 0.28/0.39/0.48 & 0.37/0.53/0.69 & 0.45/0.66/0.81 & 0.53/0.77/0.86 & 0.59/0.82/0.89 & 0.64/0.86/0.92 \\ \hline
Image 
& 64 & 0.78/0.88/0.92 & 0.83/0.92/0.95 & 0.87/0.94/0.97 & 0.90/0.95/0.98 & 0.92/0.96/0.99 & 0.94/0.98/0.99 \\ \hline
\end{tabular}}
\caption{\textbf{EEGClip.} Results of EEGClip network finetuned on EEGCVPR40 filter dataset (5-95 Hz) \cite{Spampinato2016DeepLH} for both EEG and image classification.}
\label{table:clip_eegg}
\vspace{-4mm}
\end{table*}
\section{Discussion}
\label{ref:discussion}

In this paper, we addressed the problem of EEG-to-image reconstruction and presented a comprehensive method to extract visual information from EEG data, synthesize images using extracted EEG features, and jointly train an EEG-Image model for tasks such as EEG-based image retrieval. We conducted experiments and ablation studies on three different datasets: EEGCVPR40 \cite{Spampinato2016DeepLH}, ThoughtViz \cite{kumar2018envisioned, tirupattur2018thoughtviz}, and Object \cite{kaneshiro2015representational}, using various architectures and loss functions.

We first discussed different strategies for feature extraction from EEG data, including supervised and self-supervised methods. We compared supervised classification methods with self-supervised/metric-based learning approaches and found that the latter yielded more discriminative and generalizable features, particularly using triplet loss. We demonstrated the same through improved k-means accuracy, t-SNE visualizations, and zero-shot classification.

Next, we explored the generation of images from EEG features using the GAN framework. For this, we proposed EEGStyleGAN-ADA, which incorporated EEG features and noise vectors to synthesize diverse and high-fidelity images. Our method outperformed previous EEG-to-image synthesis networks, with $62.9\%$ and $36.13\%$ inception score improvement on the EEGCVPR40 \cite{Spampinato2016DeepLH} dataset and Thoughtviz \cite{kumar2018envisioned, tirupattur2018thoughtviz} dataset, which is better than state-of-the-art performance using GAN. We have shown that achieving the photorealistic effect is not trivial with the help of class-based conditioning \ref{sec:imagesynthesis}.

Furthermore, we investigated joint representation space learning for EEG and image using the proposed EEGClip framework. We achieved significant improvements in joint representation learning by freezing the weights of a pre-trained image encoder and training an EEG feature encoder using CLIP-based loss. Upon examining the effects of batch size and epoch numbers, we observed a direct correlation between increased batch size and enhanced performance, peaking at a batch size of $64$ and $2048$ epochs, yielding scores of $79\%$ and $95\%$ for top@$1$ for EEG and image, respectively. However, extending the epoch count beyond this point showed no significant improvement. EEGClip has shown $5.96\%$ improvement over the previous state-of-the-art joint representation learning method.

\textbf{Limitations.} The proposed work has a few limitations. 1) Although we have used the same architecture for EEG feature extraction across all datasets, it is still an open problem to achieve SOTA performance using a single architecture. 2) In EEG-based image synthesis, we outperform the previous methods. Still, the quality of images in a limited dataset regime can be improved with better GAN training strategies, and we can further utilize the same for EEG-based image reconstruction.
\vspace{-1mm}
\section{Conclusion}
\label{ref:conclusion}


In this study, our primary objective was to enhance the quality of image synthesis from EEG signals. To achieve this, we introduced EEGStyleGAN-ADA, a framework capable of leveraging small and large-sized EEG datasets to generate high-resolution images $(128 \times 128)$ directly from EEG signals that outperform the previous state-of-the-art.

In addition to image synthesis, we proposed a joint representation learning framework that bridges the gap between EEG and image representation. By combining the power of both modalities, we obtained a richer and more comprehensive representation, enabling us to perform image retrieval tasks using EEG signals. Our exhaustive experiments substantiated the effectiveness of this joint representation learning approach and showcased its potential in real-world applications. Furthermore, we have also shown the triplet loss-based feature extraction method's zero-shot classification capability.

Our future endeavors will further improve EEG-based image synthesis, explore novel techniques, and refine existing frameworks. Additionally, we aim to expand our investigation to encompass the emerging field of EEG-based video analysis, leveraging temporal dynamics to generate and analyze visual content.

{\small
\bibliographystyle{ieee_fullname}
\bibliography{egbib}
}


\begin{figure*}[!t]
\centering
\begin{minipage}[b]{1.0\linewidth}
    \includegraphics[width=0.95\linewidth]{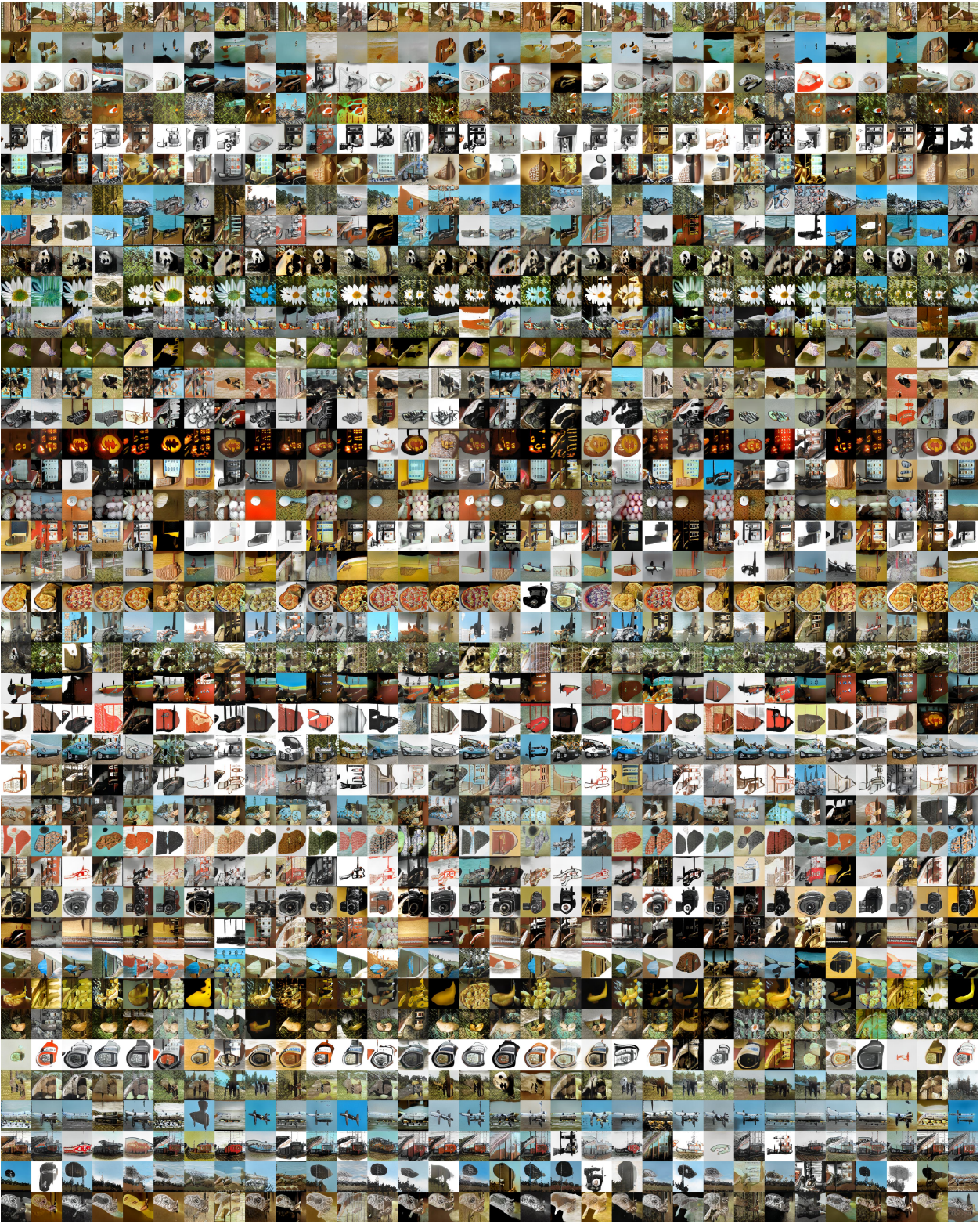}
\end{minipage}
\caption{\textbf{EEG to Image.} Images generated from EEGStyleGAN-ADA using EEG signals for all $40$ classes that show diversity and fidelity, where each image is generated with different EEG signals across different classes, EEGCVPR40 dataset \cite{Spampinato2016DeepLH}.}
\label{fig:large_genratedimages}
\end{figure*}

\begin{figure*}[!t]
\begin{minipage}[b]{1.0\linewidth}
    \centering
    \begin{subfigure}[t]{1.0\linewidth}
        \includegraphics[width=0.95\linewidth]{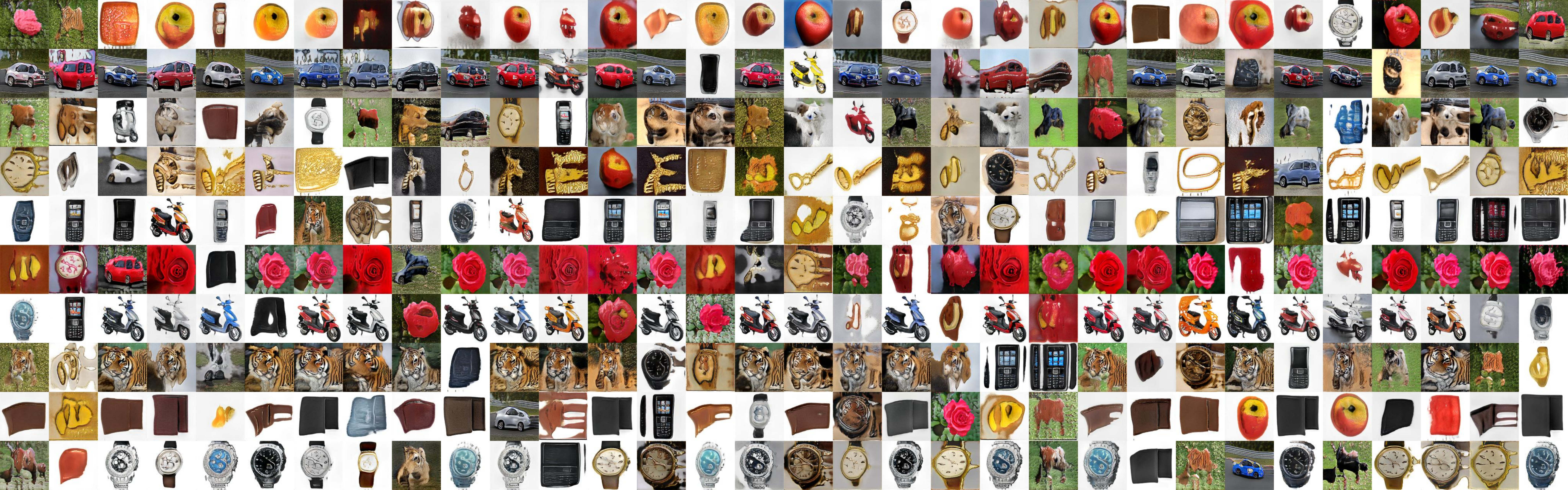}
    \end{subfigure}
    \vspace{0.1mm}
\end{minipage}
\begin{minipage}[b]{1.0\linewidth}
    \centering
    \begin{subfigure}[t]{1.0\linewidth}
        \includegraphics[width=0.95\linewidth]{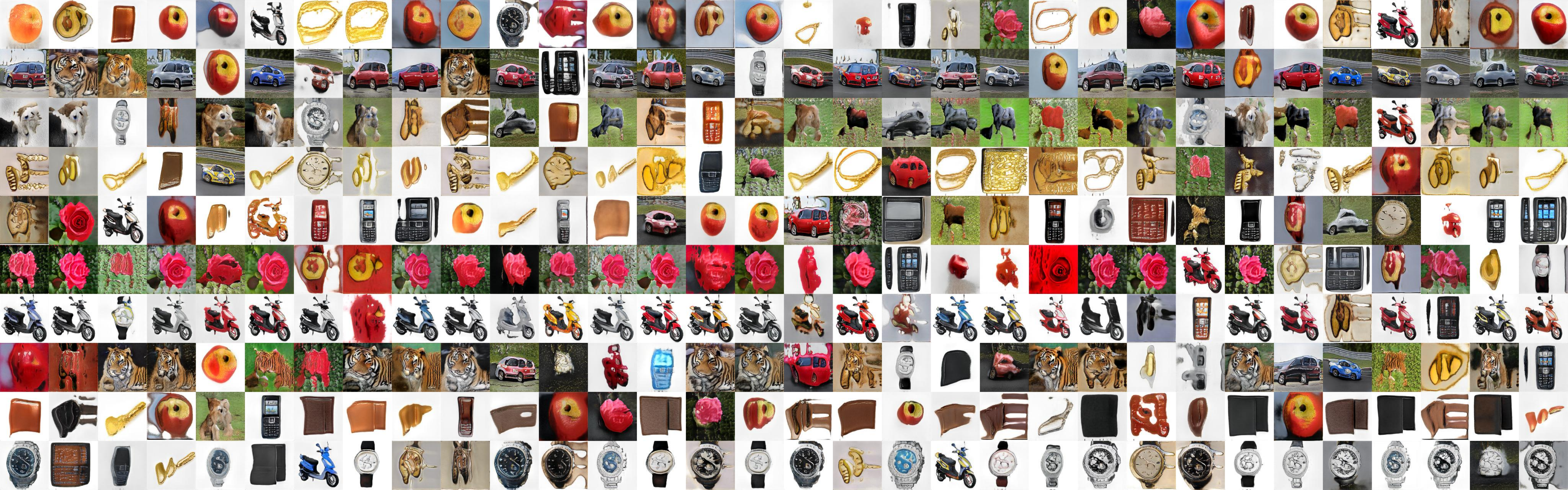}
    \end{subfigure}
\end{minipage}
\caption{\textbf{EEG to Image.} Sample images generated (top and bottom row) from EEGStyleGAN-ADA using EEG signals where each image is generated with different EEG signals across different classes, ThoughtViz dataset \cite{kumar2018envisioned, tirupattur2018thoughtviz}. There is some overlapping between classes of generated samples due to the overlapping representation learned from a small EEG dataset, which is illustrated in the main paper.}
\label{fig:genratedimages}
\end{figure*}

\end{document}